\documentclass[10pt,conference]{IEEEtran}
\usepackage[utf8]{inputenc}
\usepackage{array}
\usepackage{multirow}
\usepackage{amsmath}
\usepackage{amssymb}
\usepackage{graphicx}

\makeatletter

\providecommand{\tabularnewline}{\\}

\IEEEoverridecommandlockouts
\UseRawInputEncoding
\usepackage{cite}
\usepackage{amsfonts}\usepackage{algorithmic}
\usepackage{textcomp}
\usepackage{xcolor}
\def\BibTeX{{\rm B\kern-.05em{\sc i\kern-.025em b}\kern-.08em
    T\kern-.1667em\lower.7ex\hbox{E}\kern-.125emX}}

\@ifundefined{showcaptionsetup}{}{%
 \PassOptionsToPackage{caption=false}{subfig}}
\usepackage{subfig}
\makeatother

\begin{document}
\title{ST-Mamba: Spatial-Temporal Mamba for Traffic Flow Estimation Recovery
using Limited Data}
\author{\IEEEauthorblockN{Doncheng Yuan\IEEEauthorrefmark{1}, Jianzhe Xue\IEEEauthorrefmark{1},
Jinshan Su\IEEEauthorrefmark{2}, Wenchao Xu\IEEEauthorrefmark{3},
and Haibo~Zhou\IEEEauthorrefmark{1}} \IEEEauthorblockA{\IEEEauthorrefmark{1}School of Electronic Science and Engineering,
Nanjing University, Nanjing, China, 210023.\\
\IEEEauthorrefmark{2}Key Laboratory of Vibration Signal Capture and
Intelligent Processing, Yili Normal University, Yining, China, 835000.\\
\IEEEauthorrefmark{3}Department of Computing, the Hong Kong Polytechnic
University.\\
Email: \{dongchengyuan,jianzhexue\}@smail.nju.edu.cn, sqsjs1968@aliyun.com,
w74xu@uwaterloo.ca,\\
and haibozhou@nju.edu.cn.} }
\maketitle
\begin{abstract}
Traffic flow estimation (TFE) is crucial for urban intelligent traffic
systems. While traditional on-road detectors are hindered by limited
coverage and high costs, cloud computing and data mining of vehicular
network data, such as driving speeds and GPS coordinates, present
a promising and cost-effective alternative. Furthermore, minimizing
data collection can significantly reduce overhead. However, limited
data can lead to inaccuracies and instability in TFE. To address this,
we introduce the spatial-temporal Mamba (ST-Mamba), a deep learning
model combining a convolutional neural network (CNN) with a Mamba
framework. ST-Mamba is designed to enhance TFE accuracy and stability
by effectively capturing the spatial-temporal patterns within traffic
flow. Our model aims to achieve results comparable to those from extensive
data sets while only utilizing minimal data. Simulations using real-world
datasets have validated our model's ability to deliver precise and
stable TFE across an urban landscape based on limited data, establishing
a cost-efficient solution for TFE.
\end{abstract}


\section{Introduction}

Traffic flow estimation (TFE) assumes paramount importance in the
operation of traffic systems, particularly in light of the escalating
volume of vehicles and the expansion of transportation networks \cite{TFE_Importance1}
\cite{TFE_Importance2}. The conventional methods of TFE, employing
on-road detectors like cameras and inductive loops, exhibit limited
coverage and substantial expenses associated with infrastructure installation
and maintenance \cite{TFE_Detectors}. Utilizing data from vehicular
network can expand the scope of TFE, while the collection and transmission
of massive data also encounter technical and economic challenges.
Considering the constraints of existing methods, it is cost-effective
to utilize limited data from vehicular network to obtain real-time
TFE by cloud computing \cite{TFP_Limited}\cite{ZCY_traffic}. However,
when the amount of traffic data decreases, the resulting TFE tends
to be inaccurate and unstable \cite{JX_TSE}. With the advancement
of artificial intelligence, employing deep learning methods to recover
TFE from limited data emerges as a feasible and promising approach.

Lately, deep learning methods have been developed to explore features
of traffic flow \cite{DL_TFP} \cite{JX_URLLC}. The convolutional
neural network (CNN) is proficient in extracting spatial features
of input through convolutional operations \cite{CNN_Review}. The
recurrent neural network (RNN) and its variants such as long short-term
memory (LSTM) are designed to capture sequential or temporal features
of input by utilizing recurrent connections \cite{RNN_LSTM}. Mamba,
an advanced neural network based on state space model (SSM), is effective
at processing sequential data and modeling temporal correlation via
selection mechanism \cite{STG-Mamba}. Moreover, some methods combine
the functions of CNN and RNN. For instance, the Conv-LSTM module is
proposed to forecast short-term traffic flow via learning spatial-temporal
dependencies \cite{TFP_Conv-LSTM}. However, existing works primarily
focus on traffic flow prediction rather than estimating real-time
traffic flow, and are provided with sufficient data instead of limited
data. To enhance estimation accuracy with limited data, there is a
need to design a specialized spatial-temporal model for traffic flow
estimation recovery.

In this paper, we investigate the TFE utilizing limited data and propose
a spatial-temporal deep learning method to recover both the accuracy
and stability of TFE. The framework of TFE with limited vehicular
network data is illustrated in Fig. \ref{fig:TFE_scene}. We divide
the city map evenly into grids, each of which represent a specific
region in the city. The data limitation in our scheme refers to the
practice of randomly sampling a subset of vehicles across the city
with equal probability of recruitment at each time interval \cite{JX_TSE}.
Following the collection of limited data, the traffic information
comprising vehicle speeds and GPS coordinates are matched to corresponding
grids. The condition of traffic flow within each grid is represented
by the average speed of vehicles over a defined period. For the recovery
of TFE, we propose a novel deep learning model, named as spatial-temporal
Mamba (ST-Mamba), to generate stable and accurate TFE via effectively
leveraging the spatial-temporal features of traffic flow. Specifically,
we employ the CNN to capture the local spatial correlation of traffic
flow, and utilize the Mamba to model the temporal correlation of each
grid in the map. The main contributions of this paper are listed as
the following: 
\begin{itemize}
\item We present a novel approach for cost-effective TFE, utilizing only
a small fraction of vehicular network data to alleviate the burden
on network communication and mitigate the expenses of traffic systems.
\item We design a spatial-temporal deep learning model named as ST-Mamba,
consisting of CNN and Mamba, to recover the TFE with limited vehicular
network data.
\item We validate the effectiveness of the proposed method through comprehensive
simulations on the real-world dataset, varying in data limitation
from 10\% to 50\%. The results demonstrate that our method can provide
accurate and stable city-wide TFE.
\end{itemize}
The rest of this paper is organized as follows. Section II demonstrates
the framework and problem analysis with limited vehicular network
data. The structure of ST-Mamba is detailed in Section III. Then,
section IV showcases the simulation performance of our proposed model.
At last, we conclude this paper in Section V.

\section{Framework and Problem Analysis}

This section consists of two parts. First, the framework for TFE with
limited network data is delineated. Then, the problem of recovering
estimation from limited network data is formulated.

\subsection{Framework for TFE with Limited Data}

\begin{figure}[!t]
\centering\includegraphics[width=3.6in]{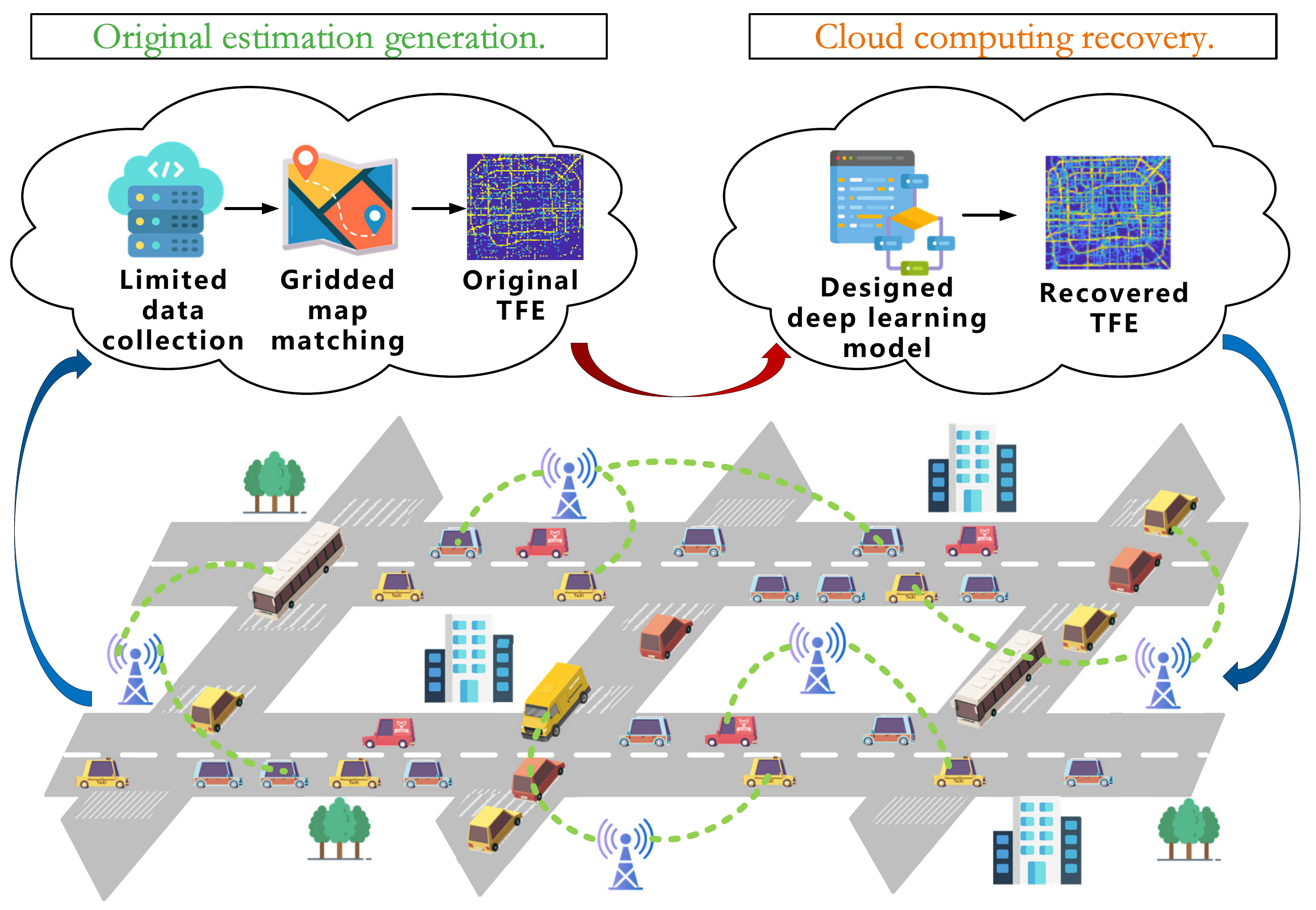}

\caption{An illustration of TFE with limited data.}
\label{fig:TFE_scene}

\vspace{0.2cm}
\end{figure}

The process of recovering TFE from limited data is illustrated in
Fig. \ref{fig:TFE_scene}. First, we divide the city map evenly into
grids. Meanwhile, the limited data of vehicle mobility information
is gathered from vehicular network by traffic data collector. Instead
of removing data from some specific regions, we acquire the limited
data by randomly selecting a small fraction of vehicles as data sources
uniformly distributed across the city, with equal probability of recruitment
for each vehicle. Following the limited data collection, we design
a matching process between the data and the grid-structured city map.
Then, the original estimation is acquired via computing the average
speeds of vehicles in each grid to construct the traffic flow image.
Lastly, the recovery algorithm is applied to generate precise TFE
based on the limited data.

As to the specific scheme of our work, we first obtain the city map
consisting of grids with equal size and number of $H${*}$W$. Corresponding
to reality, each grid in the map represents a particular region of
the city. We construct the grid-structured city map as,

\begin{equation}
\mathrm{M}=\begin{bmatrix}\mathrm{m}^{1,1} & \cdots & \mathrm{m}^{1,W}\\
\vdots & \mathrm{m}^{\text{h,w}} & \vdots\\
\mathrm{m}^{H,1} & \cdots & \mathrm{m}^{H,W}
\end{bmatrix},
\end{equation}
where $\mathrm{m}^{h,w}$ indicates the grid of map which is located
at $(h,w)$, $H$ and $W$ can be viewed as the height and width of
the map. 

With the construction of grid-structured city map, the vehicular network
data are matched to respective grids based on the GPS coordinates
of vehicles. We first obtain the vehicle mobility dataset $V_{t}$
with sufficient data, which is collected at time slot $t$ under ideal
circumstances. Subsequently, with the grid matching outcomes of $V_{t}$
and $\mathrm{M}$, the ideal estimation of traffic flow is calculated
as, 

\begin{equation}
z_{t}^{h,w}=\mathrm{Mean}(v^{h,w}|v^{h,w}\in V_{t}),
\end{equation}
\begin{equation}
\boldsymbol{Z}_{t}=\begin{bmatrix}z_{t}^{1,1} & ... & z_{t}^{1,W}\\
\vdots & z_{t}^{\text{h,w}} & \vdots\\
z_{t}^{H,1} & ... & z_{t}^{H,W}
\end{bmatrix},
\end{equation}
where the mobility information of vehicles in the grid located at
$(h,w)$ is described as $v^{h,w}$, $z_{t}^{h,w}$ is the average
speed of vehicles obtained from sufficient data in the grid located
at $(h,w)$ at time slot $t$, $Z_{t}$ represents the ideal estimation
of traffic flow at time slot $t$. 

Then, the acquisition of TFE from limited data is conducted as follows.
We denote the limited vehicle mobility information gathered at time
slot $t$ as $V_{t}^{\mathit{'}}$, which is obtained through random
sampling from the sufficient dataset $V_{t}$, i.e., $V_{t}^{'}\subseteq V_{t}$.
This sampling approach assigns equal selection probability to each
piece of vehicle mobility information. With a similar calculating
and matching process, the original estimation of traffic flow at time
slot $t$ is denoted as, 
\begin{equation}
x_{t}^{h,w}=\mathrm{Mean}(v^{h,w}|v^{h,w}\in V_{t}^{\mathit{'}}),
\end{equation}
\begin{equation}
\boldsymbol{X}_{t}=\begin{bmatrix}x_{t}^{1,1} & ... & x_{t}^{1,W}\\
\vdots & x_{t}^{\text{h,w}} & \vdots\\
x_{t}^{H,1} & ... & x_{t}^{H,W}
\end{bmatrix},
\end{equation}
where $x_{t}^{h,w}$ is the average speed of vehicles obtained from
limited data in the grid located at $(h,w)$ at time slot $t$, $X_{t}$
represents the city-wide original estimation of traffic flow at time
slot $t$. Note that both $X_{t}$ and $Z_{t}$ can be perceived as
images with a resolution of $H*W$ and four channels, where the amount
of channels is decided by the types of driving directions defined
in our work, i.e., east, south, west, and north.

Finally, we apply the specifically designed deep learning model to
recover the original estimation $X_{t}$. The ideal estimation $Z_{t}$
not only serves as the ground truth label for the learning of the
model, but also is the target data that the model expects to output.
Following the training and inference of the model, we obtain the recovered
TFE from limited data, which is reliable and cost-effective.

\subsection{Problem of Traffic Flow Estimation}

Based on the limited data, we can obtain the original estimation of
traffic flow approximating the ideal one. However, the original estimation
exhibits a considerable degree of inaccuracy and instability regardless
of the similarity. In other words, we found that there exits abundant
noise in each grid of original TFE $\boldsymbol{X}_{t}$, which fails
to serve as the definitive estimation for the applications of urban
transportation systems. Therefore, the task of real-time TFE with
limited network data is conceptualized as a grid-structured data recovery
problem. 

We set the input of the recovery model to be the historical and current
estimations of traffic flow derived from limited data, represented
as $[\boldsymbol{X}_{t-L+1},...,\boldsymbol{X}_{t-1},\boldsymbol{X}_{t}]$.
The objective is to reform an estimation of current traffic flow with
high accuracy and stability, mirroring the ideal estimation $\boldsymbol{Z}_{t}$
derived from sufficient data. Therefore, the crux of the TFE problem
lies in ascertaining the mapping function $\mathcal{F}(\cdot)$,

\begin{equation}
\boldsymbol{Z}_{t}=\mathcal{F}([\boldsymbol{X}_{t-L+1},...,\boldsymbol{X}_{t-1},\boldsymbol{X}_{t}]),
\end{equation}
which aims to fully explore the spatial-temporal dependencies of traffic
flow, mitigate the noise resulting from the data limitation, and transform
the original TFE derived from limited data into a precise estimation
of current traffic flow.

\section{ST-Mamba for Traffic Flow Recovery}

In this section, we present the ST-Mamba designed to recover accurate
and stable TFE from limited network data by exploring and aggregating
the features of traffic flow. The network structures employed for
capturing the spatial and temporal correlations by the CNN and Mamba
are discussed. Subsequently, we provide an overview of the architecture
belonging to the ST-Mamba.

\subsection{Spatial Correlation Modeling}

In our framework, the estimation of traffic flow can be viewed as
an image with a resolution of $H*W$ and four channels. The spatial
correlation of traffic flow is inherent in the grid structure. Given
that the condition of traffic flow in one region is impacted by those
in adjacent regions while the influence wanes with increasing spatial
distance, the aggregation of spatial information from neighboring
regions is imperative for accurate TFE. CNN is a classic and concise
neural network suitable for processing grid-structured data, the principle
of which is convolving input data with learnable filters to detect
local patterns \cite{CNN_Review}. In recognition of the features
of traffic flow, we employ CNN for capturing local spatial correlation
of traffic flow from the entire traffic flow image. 

We first conduct convolutional operations by CNN separately on four
channels of the image, capturing spatial correlation of each driving
direction. The working process of the CNN in our model can be denoted
as,

\begin{equation}
X_{n}^{\mathit{'}}=\mathrm{CNN_{e}}(X_{n}),
\end{equation}
where subscript $n$ ranges from $t-L+1$ to $t$, $X_{n}\in\mathbb{R}^{4\times H\times W}$
is the original TFE derived from limited data at time slot $n$, $X_{n}^{\mathit{'}}\in\mathbb{R}^{4\times H\times W}$
contains the spatial correlation captured by CNN. Then, we utilize
a fully connected (FC) layer to lift data dimension and integrate
spatial correlations of different driving directions. The FC layer
is operated as,

\begin{equation}
X_{n}^{\mathit{''}}=\mathrm{FC_{e}}(X_{n}^{\mathit{'}}),
\end{equation}
where $X_{n}^{\mathit{''}}\in\mathbb{R}^{K\times H\times W}$ is the
high-dimensional feature representation of $X_{n}^{\mathit{'}}$ in
latent space. Subsequently, one reshape layer is constructed to serialize
the format of traffic data. The sequential traffic data of each grid
can be obtained as,

\begin{equation}
I^{\mathit{h,w}}=\mathrm{Concat}(x_{n}^{h,w}|n=t-L+1,...t-1,t),
\end{equation}
where $\mathrm{Concat}(\cdot)$ is the concatenating operation, $x_{n}$
is the spatial feature representation in one grid of $X_{n}^{\mathit{''}}$,
$I\in\mathbb{R}^{K\times L}$ is the sequence comprising spatial correlation
of traffic flow in one grid of $X_{n}^{\mathit{''}}$.

\begin{figure}[!t]
\centering\includegraphics[width=3.4in]{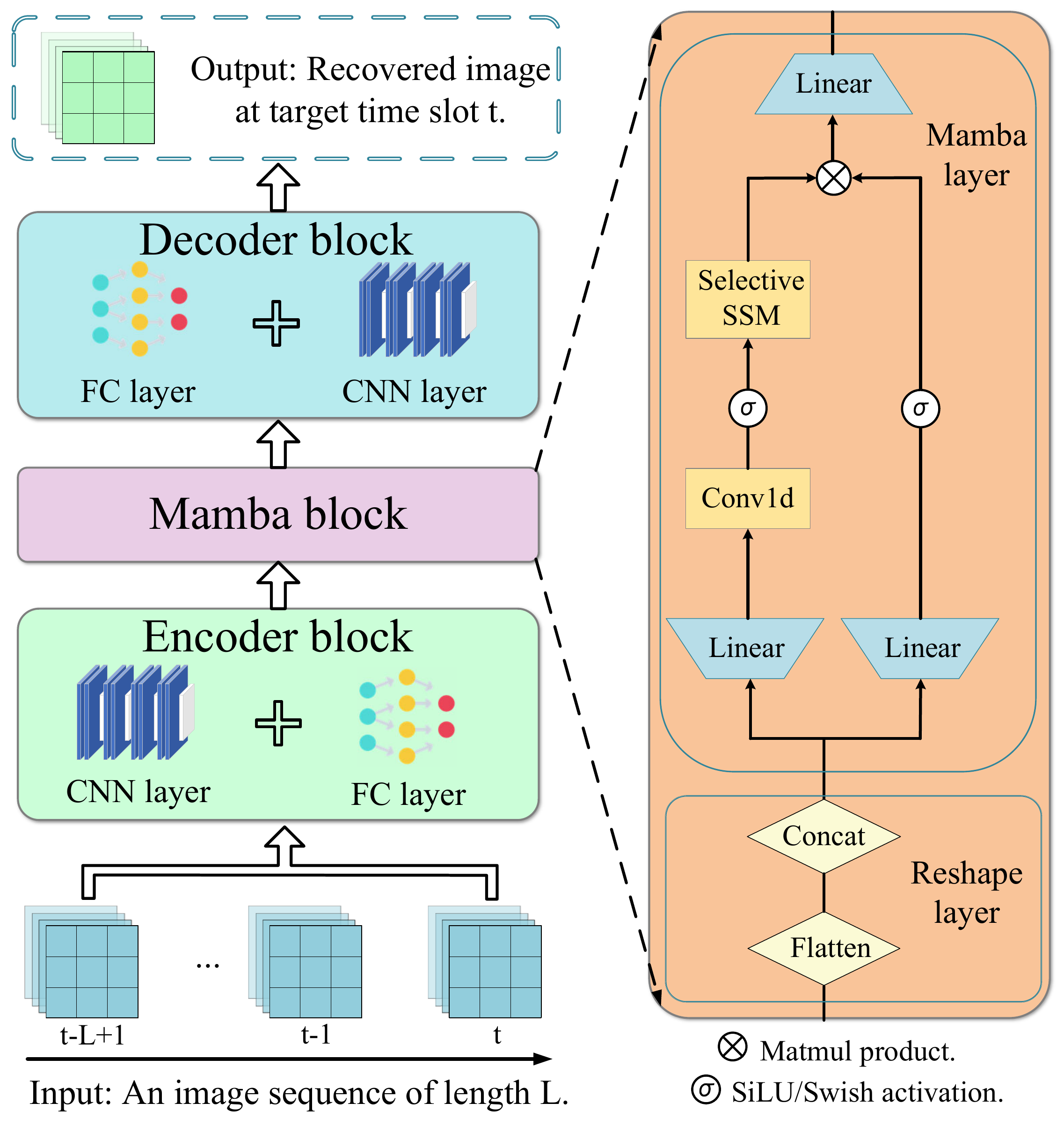}

\caption{ST-Mamba framework.}
\label{fig:Model_Framework}

\vspace{0.2cm}
\end{figure}

\subsection{Temporal Correlation Modeling}

The temporal correlation of traffic flow is hidden in the sequential
structure. The original TFE at time slot $t$, combined with those
at preceding $(L-1)$ time slots, are utilized to construct sequential
traffic data for recovering TFE at current time slot $t$. Mamba is
an advanced neural network especially effective at processing sequential
data and modeling temporal correlation. Integrating the selection
mechanism which associates the parameters of SSM with the input, Mamba
is able to focus on relevant inputs and ignore irrelevant ones, executing
linear-time processing of sequences while maintaining high performance
\cite{Mamba}. It is appropriate to employ Mamba for learning the
complex temporal correlation of traffic flow.

We aim to explore temporal correlation in each grid of the original
TFE, which directs us to set sequence $I\in\mathbb{R}^{K\times L}$
as the input of Mamba. As the pre-foundation of Mamba, we first define
the continuous-time SSM to describe the state representation of the
sequence at each time step and predict the next state based on the
input, with $\mathrm{\mathbf{A}}$ as the evolution parameter and
$\mathbf{B}$, $\mathbf{C}$ as the projection parameters. The continuous-time
SSM is formulated as,

\begin{equation}
H^{\mathit{'}}(t)=\mathbf{A}H(t)+\mathbf{B}I(t),
\end{equation}
\begin{equation}
O(t)=\mathbf{C}H(t),
\end{equation}
where $H(t)$ is the implicit hidden state at time slot $t$, $O(t)$
is the output of the SSM and is obtained via the projection of the
latest latent state.

Then, we construct Mamba with four matrix format parameters $\mathbf{\Delta}$,
$\mathrm{\mathbf{A}}$, $\mathrm{\mathbf{B}}$, $\mathrm{\mathbf{C}}$,
which is the discrete version of the continuous-time SSM. The transformation
method for discretization in Mamba is zero-order hold, which can be
written as,

\begin{equation}
\mathbf{\overline{A}}=\mathrm{exp}(\mathbf{\Delta}\mathbf{A}),
\end{equation}
\begin{equation}
\mathbf{\overline{\mathbf{B}}}=(\mathbf{\Delta}\mathbf{A})^{-1}(\mathrm{exp}(\mathbf{\Delta}\mathbf{A})-\mathbf{E})\cdot\mathbf{\Delta}\mathbf{B},
\end{equation}
where $\text{(\ensuremath{\cdot})}^{-1}$ represents the inverse matrix
operation, $\mathbf{E}$ stands for the identity matrix, $\mathbf{\Delta}$
is the timescale parameter targeted at transforming the continuous
parameters $\mathbf{A}$, $\mathbf{B}$ to discrete ones $\mathbf{\overline{A}}$,
$\mathbf{\overline{B}}$.

Subsequently, we realize the selection mechanism of Mamba by configuring
the weight matrix $\mathrm{\mathbf{B}}$, $\mathrm{\mathbf{C}}$,
$\mathbf{\Delta}$ to be input-dependent, which filters out useless
information, compresses input sequence selectively into the hidden
state, and handles contextual information effectively. Th selection
mechanism is formulated as,

\begin{equation}
\mathbf{B}=\mathrm{Linear}_{N}(I),
\end{equation}
\begin{equation}
\mathbf{C}=\mathrm{Linear}_{N}(I),
\end{equation}
\begin{equation}
\mathbf{\Delta}=\mathrm{softplus}(\mathrm{Broadcast}_{K}(\mathrm{Linear}_{1}(I))+\mathbf{D}),
\end{equation}
where $\mathbf{\mathrm{Linear}}_{d}$ is a parameterized projection
to dimension $d$, $N$ is the dimension of hidden state, $\mathrm{Broadcast}_{K}$
is targeted at broadcasting one-dimensional data into a $K$-dimensional
space, $\mathbf{D}$ is a constant weight matrix with dimension $K$,
$\mathrm{softplus}$ ensures numerical stability as an activation
function. Note that $\mathbf{A}$ comprises constant parameters for
simplicity, while $\mathbf{\overline{A}}$ is input-dependent via
the process of discretization by $\mathbf{\Delta}$. Meanwhile, although
$\mathbf{B}$ and $\mathbf{C}$ are computed in a similar way, they
are independent sets of parameters and represent different dynamics
in Mamba.

Following the discretization of parameters and the implementation
of selection mechanism, we conduct the computation of global convolution
to obtain the output of Mamba as, 

\begin{equation}
\mathbf{\overline{S}}=(\mathbf{C\overline{B}},\mathbf{C\overline{A}\overline{B}},...,\mathbf{C}\mathbf{\overline{A}}^{L-1}\mathbf{\overline{B}}),
\end{equation}
\begin{equation}
O=I\ast\mathbf{\overline{S}},
\end{equation}
where $\mathbf{\overline{S}}$ is a structured convolutional kernel
for implementing the selective scan algorithm to speed up the model,
$O\in\mathbb{R}^{K\times L}$ is the final output of Mamba which contains
spatial and temporal correlations of traffic flow in each grid. 

\subsection{Outcome Decoding}

The final outcome is decoded from the output of Mamba. The decoder
block consists of one FC layer and one CNN layer. In the FC layer,
we combine the sequential traffic data $O$ to form an image format
with dimension $\mathbb{R}^{(K\times L)\times H\times W}$. Meanwhile,
we reduce data dimension to aggregate spatial-temporal correlations
previously modeled. The FC layer in decoder block is operated as,

\begin{equation}
G=\mathrm{FC_{d}}(O),
\end{equation}
where $G\in\mathbb{R}^{4\times H\times W}$ can be viewed as the image
of TFE containing spatial and temporal correlations of traffic flow.
Then, we conduct convolutional operations in the CNN layer to decode
the outcome as, 

\begin{equation}
\boldsymbol{Y}=\mathrm{CNN_{d}}(G),
\end{equation}
where $\boldsymbol{Y}\in\mathbb{R}^{4\times H\times W}$ is the final
outcome of the model and can be perceived as the recovered TFE approximating
the ideal estimation $\boldsymbol{Z}_{t}$.

\subsection{ST-Mamba Model}

The main architecture of ST-Mamba model is illustrated in Fig. \ref{fig:Model_Framework}.
We specifically design three parts for ST-Mamba, which are the encoder
block, the Mamba block, and the decoder block, respectively. 

In specific, the encoder block consists of one CNN layer and one FC
layer in turn. Given that vehicles driving in different directions
have minimal influence on each other, four CNNs with identical parameters
are employed in the CNN layer to independently perform convolutional
operations in each channel of input, extracting spatial correlation
of traffic flow under four distinct driving directions. Then, we utilize
one FC layer to lift the dimension of input data to explore much more
detailed information and exchange spatial features of different driving
directions. For the Mamba block, we stack one reshape layer before
the Mamba layer, serializing the input so that Mamba can process the
limited data of traffic flow effectively. Then, the generated sequences
are fed into the Mamba layer in parallel to explore the temporal correlation
in each gird of the original traffic flow estimations. For the decoder
block which is similar to encoder one, the role of the FC layer is
to aggregate spatial-temporal correlations captured previously and
project the processed data into the image format. Then, convolutional
operations are done by the CNN layer to decode the final outcome while
form a formal symmetry of the model. Finally, ST-Mamba exports the
recovered traffic flow estimation at time slot $\mathrm{\mathit{t}}$
from limited data, with high accuracy and stability which can meet
the demands of traffic transportation systems.

\begin{figure*}[!t]
\centering\subfloat{\includegraphics[width=7.15in]{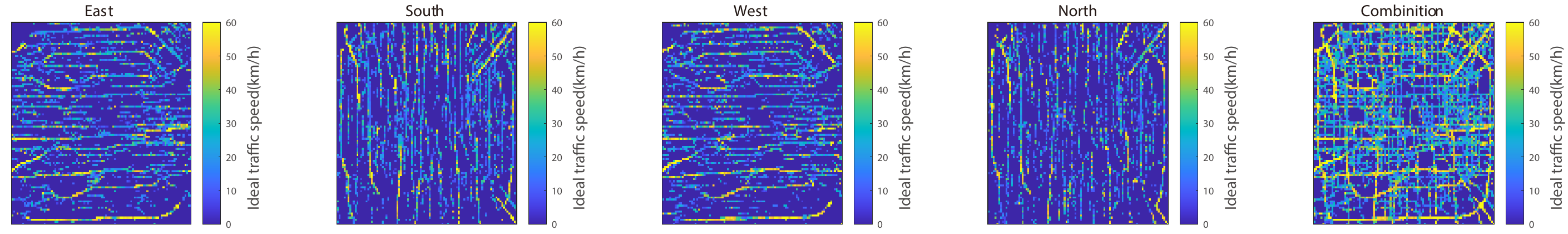}}

\caption{The ideal TFE on Friday 5:00 PM.}
\label{fig:grid_ideal}
\end{figure*}

\begin{figure*}[!t]
\centering\subfloat{\includegraphics[width=7.15in]{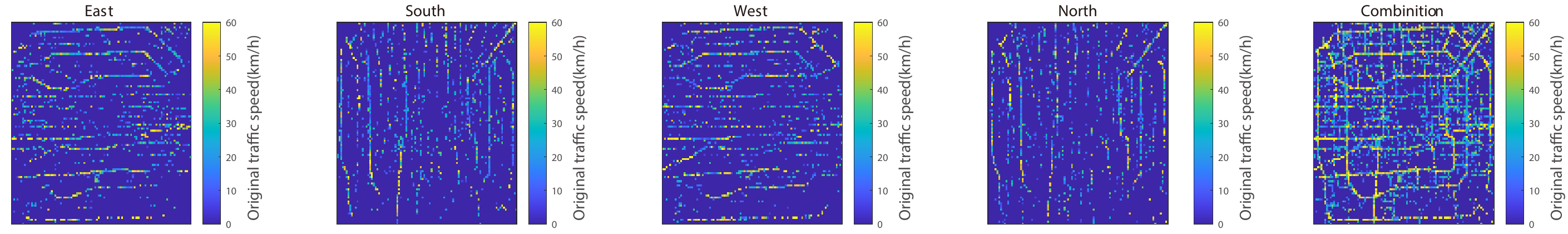}}

\caption{The original TFE on Friday 5:00 PM at 10\% limitation.}
\label{fig:grid_origin}
\end{figure*}

\begin{figure*}[!t]
\centering\subfloat{\includegraphics[width=7.15in]{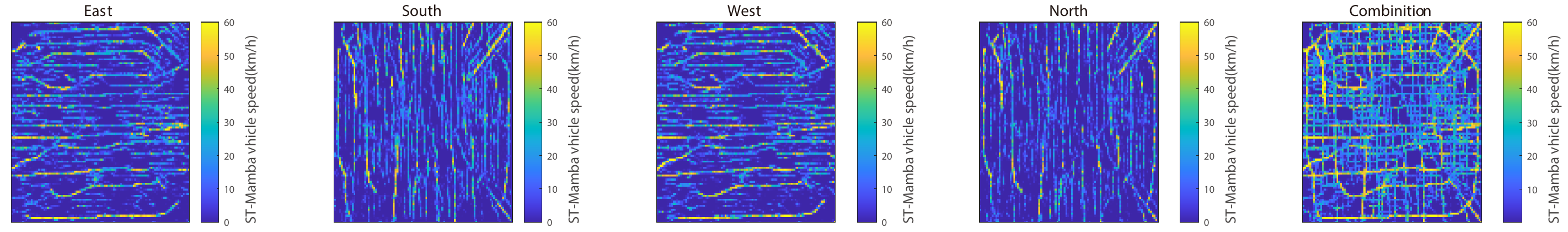}}

\caption{The recovered TFE on Friday 5:00 PM at 10\% limitation.}
\label{fig:grid_recover}
\end{figure*}

\section{Simulation and Results}

In this section, we assess our method utilizing the real traffic data
from Beijing. We begin by the experimental settings comprising the
dataset, baselines, and evaluation metrics. Then, we present and analyze
the estimation performance.

\subsection{Experimental Setting}

The simulation utilizes the real traffic data sourced from the Fourth
Ring Road in Beijing. Our dataset includes 6 days in 2012, and the
time period for TFE spans from 7:30 AM to 10:30 PM \cite{JX_TSE}.
The resolution of TFE image is 100$\times$100. The limited data is
obtained by random sampling from sufficient data, where the limitation
degree indicates the proportion of sampled data. Our approach is compared
with the following four baselines:
\begin{itemize}
\item Original: The original estimation derived from limited data, and the
input of deep learning models. 
\item CNN: A convolutional neural network designed to capture spatial correlation
by convolutional operations \cite{CNN_Review}. 
\item PredRNN: A recurrent neural network proposed to capture spatial-temporal
dependencies for coherent future frame forecasting \cite{Predrnn}. 
\item SimVP: A simple yet powerful architecture for video prediction, which
makes the most of CNN and its variants to explore spatial-temporal
correlations \cite{SimVP}. 
\end{itemize}

Meanwhile, we select three metrics to evaluate the performance of
models. Root mean squared error (RMSE) assesses the recovery accuracy
by penalizing larger errors more heavily. Improved percentage (IP)
demonstrates the recovery improvement brought by the model. Mean absolute
error (MAE) calculates the average absolute difference between the
recovered and ideal estimation.

\begin{figure}[!t]
\centering\includegraphics[width=3.5in]{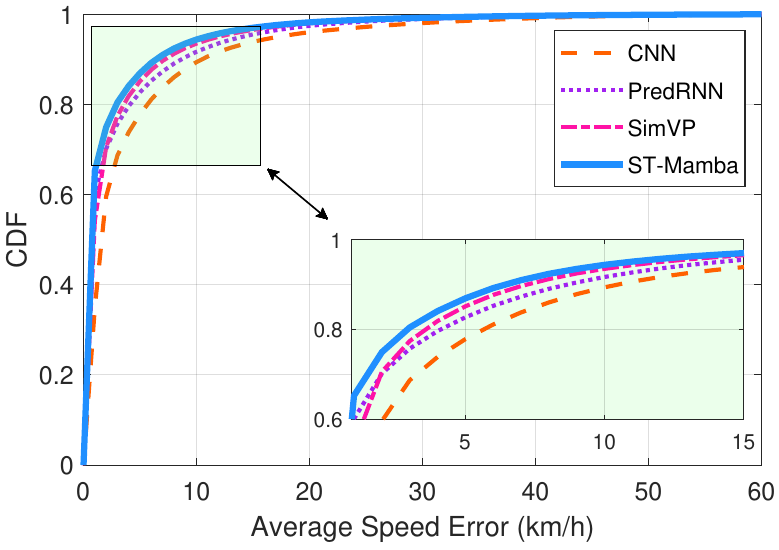}

\caption{The CDFs of estimation error on Sunday.}
\label{fig:cdf}
\end{figure}

\subsection{Performance Analysis}

\begin{table}
\caption{The estimation recovery comparison.}

\centering{} %
\begin{tabular}{|c|c|>{\centering}p{0.55in}>{\centering}p{0.55in}>{\centering}p{0.55in}|}
\hline 
\multirow{2}{*}{Limitation} & \multirow{2}{*}{Model} & \multicolumn{3}{c|}{Metric}\tabularnewline
\cline{3-5} \cline{4-5} \cline{5-5} 
 &  & RMSE & IP & MAE\tabularnewline
\hline 
\multirow{5}{*}{10\%} & Original & 13.269 & {*} & 5.379\tabularnewline
 & CNN & 9.955 & 24.979\% & 5.359\tabularnewline
 & PredRNN & 8.067 & 39.207\% & 3.965\tabularnewline
 & SimVP & 8.224 & 40.864\% & 4.561\tabularnewline
 & ST-Mamba & $\mathbf{7.504}$ & $\mathbf{43.449\%}$ & $\mathbf{3.531}$\tabularnewline
\hline 
\multirow{5}{*}{20\%} & Original & 11.072 & {*} & 4.073\tabularnewline
 & CNN & 8.786 & 20.638\% & 5.124\tabularnewline
 & PredRNN & 6.966 & 37.082\% & 3.231\tabularnewline
 & SimVP & 7.049 & 38.415\% & 3.680\tabularnewline
 & ST-Mamba & $\mathbf{6.379}$ & $\mathbf{42.386\%}$ & $\mathbf{2.851}$\tabularnewline
\hline 
\multirow{5}{*}{30\%} & Original & 9.481 & {*} & 3.204\tabularnewline
 & CNN & 7.957 & 16.065\% & 4.067\tabularnewline
 & PredRNN & 6.443 & 32.043\% & 2.843\tabularnewline
 & SimVP & 6.126 & 37.042\% & 3.001\tabularnewline
 & ST-Mamba & $\mathbf{5.516}$ & $\mathbf{41.813\%}$ & $\mathbf{2.245}$\tabularnewline
\hline 
\multirow{5}{*}{40\%} & Original & 8.187 & {*} & 2.547\tabularnewline
 & CNN & 7.134 & 12.854\% & 3.529\tabularnewline
 & PredRNN & 5.967 & 27.118\% & 2.615\tabularnewline
 & SimVP & 5.341 & 36.183\% & 2.537\tabularnewline
 & ST-Mamba & $\mathbf{4.857}$ & $\mathbf{40.670\%}$ & $\mathbf{1.876}$\tabularnewline
\hline 
\multirow{5}{*}{50\%} & Original & 7.056 & {*} & 2.015\tabularnewline
 & CNN & 6.631 & 6.008\% & 3.077\tabularnewline
 & PredRNN & 5.817 & 17.549\% & 2.472\tabularnewline
 & SimVP & 4.703 & 34.664\% & 1.989\tabularnewline
 & ST-Mamba & $\mathbf{4.288}$ & $\mathbf{39.220\%}$ & $\mathbf{1.540}$\tabularnewline
\hline 
\end{tabular}\label{tab:results}
\end{table}

Table \ref{tab:results} presents a comparative analysis of recovery
performance by the ST-Mamba against other baselines. Models are evaluated
across five degrees of data limitation from 10\% to 50\%. ST-Mamba
demonstrates superior recovery capability compared to the other models,
consistently attaining the lowest RMSE and MAE, alongside the highest
IP. These results indicate that the traffic flow estimations recovered
by ST-Mamba have the least error. In addition, considering the significant
errors associated with original TFE, the errors of ST-Mamba are deemed
acceptable for practical applications.

As to the recovery performance within space domain, Fig. \ref{fig:grid_ideal},
Fig. \ref{fig:grid_origin}, and Fig. \ref{fig:grid_recover} show
the TFE under different conditions. Fig. \ref{fig:grid_ideal} is
the ideal TFE, calculating the average vehicle speeds from 100\% data
and clearly demonstrating the framework of Fourth Ring Road. Fig.
\ref{fig:grid_origin} is the original TFE at the data limitation
degree of 10\%, where considerable number of data missing and errors
can be witnessed. Fig. \ref{fig:grid_recover} is the recovered TFE
obtained by ST-Mamba and is similar to the ideal one in terms of both
grid color and image structure, proving excellent recovery performance
of ST-Mamba in space domain.

The recovery performance within time domain is illustrated in Fig.
\ref{fig:cdf}, where we display the cumulative distribution function
(CDF) of the estimation error per minute from 7:30 to 22:30 on Sunday
at limitation degree of 30\%. It can be observed the CDF of ST-Mamba
converges the fastest, which means the errors of ST-Mamba are concentrated
on small values and the error range is relatively narrow compared
to baselines. In other words, ST-Mamba is able to provide reliable
recovery for TFE throughout a day.

In general, ST-Mamba is able to capture spatial-temporal correlations
of traffic flow effectively, providing accurate and stable TFE for
the entire city on various days.

\section{Conclusion}

In this paper, we studied the framework for TFE utilizing limited
data from the vehicular network. We analyzed the TFE recovery problem
caused by the limitation of traffic data availability. To address
this issue, we designed a spatial-temporal deep learning model named
as ST-Mamba to improve the estimation accuracy and stability. Comprehensive
simulation results based on the real-world dataset have validated
the effectiveness of ST-Mamba in processing sequential grid-structured
data. Our approach provides a cost-effective solution for TFE in scenarios
with limited data. For future work, we intend to explore the traffic
flow prediction based on limited data.

\section*{Acknowledgment}

This work is supported in part by the National Natural Science Foundation
of China under Grant 623B2052, 62271244, the Natural Science Fund
for Distinguished Young Scholars of Jiangsu Province under Grant BK20220067.

\bibliographystyle{IEEEtran}
\bibliography{C_Reference}

\end{document}